\documentclass{llncs} 
\usepackage{amsmath,amssymb,amsfonts} 
\usepackage{graphics}                 
\usepackage{color}                    
\usepackage{hyperref}                 

\usepackage{enumitem}
\usepackage{graphicx}
\graphicspath{{Figs/}}

\def\RealSet{\mbox{$I\! \! R$}}

\title{Multitask Kernel-based Learning with First-Order Logic Constraints}
\titlerunning{FOL Constraints in Kernel Learning}

\author{Michelangelo Diligenti \and Marco Gori \and Marco Maggini \and Leonardo Rigutini}
\authorrunning{M. Diligenti et al.}
\tocauthor{Michelangelo Diligenti, Marco Gori, Marco Maggini, Leonardo Rigutini}

\institute{Dipartimento di Ingegneria dell'Informazione \\
Universit\`a di Siena, Italy \\
\email{\{michi,marco,maggini,rigutini\}@dii.unisi.it}
}

\begin{document}

\maketitle

\begin{abstract}
In this paper we propose a general framework to integrate supervised and unsupervised examples with background knowledge expressed by a collection of first-order logic clauses into kernel machines. 
In particular, we consider a multi-task learning scheme where multiple predicates defined on a set of objects are to be jointly learned from examples, enforcing a set of FOL constraints on the admissible configurations of their values.
The predicates are defined on the feature spaces, in which the input objects are represented, and can be either known a priori or approximated by an appropriate kernel-based learner. 
A general approach is presented to convert the FOL clauses into a continuous implementation that can deal with the outputs computed by the kernel-based predicates. 
The learning problem is formulated as a semi-supervised task that requires the optimization in the primal of a loss function that combines a fitting loss measure on the supervised examples, a regularization term, and a penalty term that enforces the constraints on both the supervised and unsupervised examples. Unfortunately, the penalty term is not convex and it can hinder the optimization process. However, it is possible to avoid poor solutions by using a two stage learning schema, in which the supervised examples are learned first and then the constraints are enforced.
\end{abstract}

\section{Introduction}
This paper proposes a general framework to inject background knowledge expressed by first-order logic clauses into the regularized fitting of supervised examples carried out by kernel machines~\cite{Scholkopf2001}.
This approach has strong connections to studies of developmental psychology, where children have been shown to iteratively generalize their knowledge, initially derived from a set of punctual examples and then properly increased by the gradual acquisition of higher level concepts ~\cite{gori2009}.
While the focus on biologically-plausible solutions has been playing a central role,  scientists have mostly neglected those human behavioral principles, and 
have not been able to bridge learning processes based on examples and high level logic representations. 
Important exceptions concerning a unified treatment of learning with prior knowledge in logic form have been proposed in the field of probabilistic inductive logic programming (see e.g.~\cite{Muggleton2005} and~\cite{Frasconi2008}).
In this paper, we consider a multi-task learning scheme where multiple predicates defined on a set of objects are to be jointly learned from examples, enforcing a set of FOL constraints on the admissible configurations of their values.
The predicates are implemented as generic functions getting as input a vector of features.
A general approach is presented to convert any set of FOL clauses into a set of constraints on real-valued functions. We proposed a semi-supervised learning approach that requires the optimization of a function composed of a  loss function, a regularization term, and a penalty term that enforces the constraints. When enforcing the constraint satisfaction term  on the supervised and unsupervised examples only, we show that a representation theorem holds  that dictates  the optimal solution of the problem as a kernel expansion over the given examples.
Unfortunately, unlike for classic kernel machines, the cost function is not guaranteed to be convex, unless for very simple FOL clauses.
While direct optimization of the cost function is hopeless, we propose a solution inspired to stage-based learning  in which we learn the supervised examples first and then we enforce the FOL constraints.
The proposed approach is  general and flexible. As an example, we show that it is related to  manifold regularization~\cite{belkin2006manifold}, which also emerge in our logic setting. The experimental results show how the background knowledge increases the classification accuracy in the context of multi-task classification problems.

This is the outline of the paper. Section \ref{sec:learning_with_constraints} introduces the proposed learning framework, while section \ref{sec:experiments} reports some experimental results. Finally, section \ref{sec:conclusions} draws some conclusions.

\section{Learning with first-order logic constraints}
\label{sec:learning_with_constraints}
We consider a learning problem in which a set of predicates is to be inferred from
examples. For the sake of simplicity, we assume that the arguments of each predicate
can take values from the same domain, which is assumed to be a vectorial feature space
$F \in \RealSet^m$. However, it is easy to extend the framework to consider a different
domain for each variable. The predicates are implemented as functions from
a Cartesian product of the feature space to a scalar real value, such that
$\pi_k : F^{n_k} \rightarrow \RealSet, k=1,\ldots,T$ being $n_k$ the grade of the predicate
(i.e. the number of arguments). The value computed by the function $\pi_k$ can
be interpreted as a continuous truth value, that is associated to the corresponding
predicate for each configuration of the input variables.

The learning task is defined in a semi-supervised scheme by assigning a set of
labeled examples for each predicate together with a set of unlabeled samples that
are drawn from the feature vector distribution. In particular, the labeled examples for the $k$-th predicate are collected in the set ${\cal L}_k = \{ \left([{\bf x}_1^i,\ldots,{\bf x}_{n_k}^i],y_k^i\right) | i=1,\ldots, \ell_k\}$
where ${\bf x}_j^i \in F, j=1,\ldots,n_k$, is the $j$-th argument of the predicate, and $y_k^i \in \{0,1\}$ is the target that encodes the truth value of the predicate for the given input configuration. The unlabeled set is
${\cal U} = \{ {\bf x}^i | {\bf x}^i\in F, \ i=1,\ldots,u\}  \,$
and it is possible to collect all the available feature vectors in the set
${\cal S} = \{{\bf x} | \exists k \ \exists i \left([{\bf x}_1,\ldots,{\bf x}_i ,\ldots, {\bf x}_{n_k}],y_k\right) \in {\cal L}_k \land {\bf x}={\bf x}_i\} \ \cup \  {\cal U} \ .
$
We assume that beside the information provided by the given labeled and unlabeled
examples, further a priori knowledge on the given task is modeled by a set of First-Order Logic (FOL) clauses
defined over the unknown predicates $\pi_k$ and a set of known predicates $\pi^K_j$,
$j=1,\ldots,T^K$. Basically the learning task requires to determine the unknown
predicates such that they provide an optimal fitting of the supervised examples 
satisfying also the a priori FOL clauses on the feature space. We assume that each unknown
predicate $\pi_k$ can be conveniently approximated in a given Reproducing Kernel Hilbert Space (RKHS)
${\cal H}_k$ by a function $f_k : F^{n_k} \rightarrow \RealSet$, such that the learning problem can be cast
as an optimization task, where the objective can be expressed as
\begin{equation}
E(\mathbf {f}) = R(\mathbf {f}) + N(\mathbf {f}) + V(\mathbf {f})
\label{eq:objective}
\end{equation}
where $\mathbf {f}=[f_1,\ldots,f_T]^\prime$ is the array of the candidate predicates. The fitting of the supervised
examples is considered by the term
$$
	R({\bf f}) = \displaystyle \sum_{k=1}^T \lambda^\pi_k \cdot\frac{1}{\left|{\cal L}_k \right| }
    \sum_{([{\bf x}_1,\ldots,{\bf x}_{n_k}],y)\in{\cal L}_k} L^e_k(f_k({\bf x}_1,\ldots,{\bf x}_{n_k}),y),
$$
where $L_k^e(z,y)$ is a loss function that measures the fitting quality of $f_k(\cdot)$ with respect to the target $y$ and $\lambda^\pi_k>0$ is the weight for the $k$-th predicate. 
Even if extending this approach to employ multi-task kernels is trivial, in this paper we consider scalar kernels that do not yield interactions amongst the different predicates, that is 
$
N({\bf f}) = \sum_{k=1}^T \lambda^r_k \cdot ||f_k||^2_{{\cal H}_k},
$
where $\lambda^r_k>0$ can be used to impose a different weight to each predicate. The last term
$V(\mathbf {f})$ enforces the assigned FOL constraints by penalizing their violation. The constraints
are assumed to hold for any valid configuration as defined by the quantified variables in the
FOL expression of each clause, however the penalty considers only the sampling yielded by
the points in ${\cal S}$. The constructive procedure to convert an FOL clause into an appropriate
penalty will be described in the following subsection.

It is easy to prove a straightforward extension of the Representer Theorem for plain kernel machines~\cite{Scholkopf2001}
that states that the solution of the optimization problem involving the objective function of
eq.~(\ref{eq:objective}) can be expressed by a kernel expansion on the given examples available
in the set ${\cal S}$. In fact, similarly to the term corresponding to the empirical risk, the penalty term to enforce the constraints only involves values of $f_k$ sampled on a subset of ${\cal S}^{n_k}$. Hence the optimal approximation for the $k$-th predicate can be written as
$$
f_k({\bf x}_1,\ldots,{\bf x}_{n_k}) = \sum_{[{\bf x}^i_1,\ldots,{\bf x}^i_{n_k}] \in  {\cal S}_k}
w_{k,i} K_k([{\bf x}^i_1,\ldots,{\bf x}^i_{n_k}],[{\bf x}_1,\ldots,{\bf x}_{n_k}])
$$
where $K_k(\cdot,\cdot)$ is the reproducing kernel associated to the space ${\cal H}_k$ and ${\cal S}_k \subseteq {\cal S}^{n_k}$
is an appropriate subset of the $n$-tuples that can be formed from the available sample points.
The representer theorem shows that it is possible to optimize eq.~(\ref{eq:objective}) in the primal by gradient heuristics~\cite{chapelle2007training}. 
The weights of the kernel expansion can be compactly organized in ${\bf w}_k = [w_{k,1},\ldots,w_{k,{|{\cal S}_k|}}]^\prime$
and, therefore, the optimization of eq.~(\ref{eq:objective}) turns out to involve the finite set of real-valued
weights ${\bf w}_k, k=1,\ldots,T$. The overall error function might not be convex anymore due to the constraint penalty term. 
However, in case of positive kernel, the strict convexity is guaranteed when restricting the learning to the regularization and empirical risk terms. 
Please note that the labeled examples and the constraints are coherent, since they represent different reinforcing expressions of the concepts to be learned. If we start applying the constraint penalty term only after having learned the supervised examples then the penalty term should be null, when restricted to the supervised portion of examples and non-null only on the unsupervised portion. Hence, the proposed learning procedure consists of two consecutive stages: \emph{\sc Labeled initialization}, in which only a regularized fitting of the supervised examples is enforced, and
\emph{\sc abstraction stage} during which we also start enforcing the constraints in the cost function.
This technique has been proven to effectively tackle non-convex learning tasks.


\subsection{Enforcing FOL clauses}
To integrate FOL clauses into the proposed learning framework, we can exploit the classic association from Boolean variables to real-valued functions by using the
\emph{t-norms} (triangular norms)~\cite{Klement2000}. A t-norm is any function
$T: [0,1] \times [0,1] \rightarrow \RealSet$, that is commutative, associative, monotonic (i.e. $y \le z \Rightarrow T(x,y) \le T(x,z)$),
and featuring a neutral element 1 (i.e. $T(x,1) = x$). A t-norm fuzzy logic is defined by its t-norm $T(x,y)$
that models the logic AND, while the negation of a variable $\lnot x$ is computed as $1-x$.
The {\emph t-conorm}, modeling the logical OR, is defined as $1 - T((1-x), (1-y))$, as a generalization of the De Morgan's law.
In the following we will consider the product t-norm $T(x,y) = x \cdot y$, but other choices are possible.

Hence, using a t-norm it is possible to define the logic operators that can be used to
implement the continuous counterpart of a FOL expression. In particular, if $e_1$ and $e_2$ are the functions
implementing two FOL expressions $E_1$ and $E_2$, eventually depending on a
set of variables, when exploiting the product t-norm, we have that $E_1 \land E_2$ is
implemented by $e_1 \cdot e_2$, $E_1 \lor E_2$ by $1-(1-e_1)\cdot(1-e_2)$, and finally
$\lnot E_1$ as $1-e_1$. The atoms in the expressions are represented by both the assigned predicates
$\pi^K_j$ and the unknown predicates $\pi_k$, whose arguments are defined by a set
of  quantifiers. When considering the continuous implementation, the predicates $\pi_k$ are
approximated by the corresponding kernel expansion $f_k$ that is not guaranteed
to yield values in the interval $[0,1]$ as required by the definition of t-norms.
Hence, we apply a squashing function $\sigma : \RealSet \rightarrow [0,1]$ to constrain their values in $[0,1]$,
such that the atoms related to the predicates $\pi_k$ are implemented as
$\sigma(f_k({\bf x}_1,\ldots,{\bf x}_{n_k}))$. In the experimental setting, we exploited the
targets $\{0,1\}$ for the $\{false,true\}$ values in the supervised examples, and we
decided to exploit the piecewise-linear squash function $\sigma(y) = \mbox{min}(1,\mbox{max}(y,0))$.

The expressions obtained by combining the predicates with logic operators contain variables corresponding to
the arguments of each predicate. Each variable can range in the feature space $F$ and,
when the same variable is shared among different arguments and/or predicates, it is
assumed that same value should be used in all its instances. The definition of the FOL
clauses, that represent the given a priori constraints, requires each variable to be properly
quantified in order to obtain an expression that should evaluate to the value $true$ for
all valid hypotheses for the unknown predicates $\pi_k$. In other words, the constraint
can be thought as a functional, depending on the functions $f_k$, that should evaluate to
$1$ only for the solutions satisfying the constraint. Hence, if we denote by $C_h$ the
constraint represented by the $h$-th FOL clause, its dependence from the functions $f_k$
can be make explicit by writing $C_h(f_1,\ldots,f_T)$. The resulting expressions for all the
clauses can be used to define the penalty term $V({\bf f})$ in eq. (\ref{eq:objective}).
We first consider how to implement the universal quantifier. If we assume that the
expression $E$ depends on the variable ${\bf x} \in F$, a clause can be defined as
$\forall \ {\bf x} \ E({\bf x})$ meaning that the proposition $E({\bf x})$ should be $true$ for
all points in the feature space. Since it is infeasible to verify the constraint on all the
feature space, we exploit the dataset ${\cal S}$ to evaluate the constraint. In particular,
we force the t-norm expression of $E({\bf x})$ to be $true$ in average by defining
the penalty
$
V_e({\bf f }) = \frac{1}{|{\cal  S}|}\sum_{{\bf x} \in {\cal  S} } (1-e({\bf x})) \ .
$
In fact, the term 1-e({\bf x}) assumes values in $[0,1]$ and is null when the
expression is verified in ${\bf  x}$. Hence, the penalty is null when the expression
is verified over all the sample points. In general, when the expression depends on a set of
variables that are universally quantified, i.e. the clause is in the form $\forall \ {\bf x}_1
\ldots \forall {\bf x}_q \ E({\bf x}_1,\ldots,{\bf x}_q)$, the penalty can be written as
$$
V_e({\bf f }) = \frac{1}{|{\cal  S}|^q}\sum_{{\bf x}_1 \in {\cal S} } \ldots \sum_{{\bf x}_q \in {\cal  S} } (1-e({\bf x}_1,\ldots,{\bf x}_q)) \ .
$$
Clearly, the complexity of the penalty is combinatorial in the cardinality of the sample set,
but we can usually exploit the fact that some correlations among the different variables
are present in the real problem. For instance, we can model the fact that some
configurations of the quantified variables are not admissible by introducing the predicate
$d({\bf x}_1,\ldots,{\bf x}_q)$ that is true only for the valid (or more probable)
configurations of the variables and by enforcing the proposition $d({\bf x}_1,\ldots,{\bf x}_q) \Rightarrow e({\bf x}_1,\ldots,{\bf x}_q)$. In fact, the proposition is equivalent to $\lnot (d({\bf x}_1,\ldots,{\bf x}_q) \land \lnot e({\bf x}_1,\ldots,{\bf x}_q))$ whose t-norm implementation is $1-d({\bf x}_1,\ldots,{\bf x}_q) \cdot (1-e({\bf x}_1,\ldots,{\bf x}_q))$. Thus the term involved in the penalty is $d({\bf x}_1,\ldots,{\bf x}_q) \cdot (1-e({\bf x}_1,\ldots,{\bf x}_q))$ that is null whenever the variable configuration is not valid.
This result is implemented by computing the sum only on the $n$-tuples observed in the dataset.

The implementation of the existential quantifier is more tricky, since it requires to verify the existence in the feature space of a point that satisfies a given proposition. Given the expression $\exists{\bf  x}~E({\bf x})$, we have no a priori general rule to find the feature vectors ${\bf x} \in F$ that make $E({\bf x})$ true for the current hypothesis of $f_k$.
Assuming that at least one of such points is available in the sample set ${\cal S}$, the existence operator can be rewritten as $\bigvee_{{\bf x}_i \in {\cal S}} E({\bf x}_i)$. This expression can be transcripted in a continuous t-norm form as: $1-\prod_{{\bf x}_i \in {\cal S}} (1-e({\bf x}_i))$. Please note that an efficient implementation of this operator needs further studies.

\section{Experimental results}
\label{sec:experiments}
This section presents an experimental analysis on some artificial benchmarks properly created to show how the FOL logic clauses can be used in the context of semi-supervised learning with plain kernel machines.
The two-stage learning algorithm described in section~\ref{sec:learning_with_constraints} is exploited in all the experiments.

A first synthetic task consists of a multi-class classification problem over four different classes: $A,B,C,D$.
The patterns for each class are assumed to be uniformly distributed over the following rectangles:
$
A = \{(x,y): 0 \le x \le 3, 0 \le y \le 3\}, B = \{(x,y): 1 \le x \le 4, 1 \le y \le 4\}, 
C = \{(x,y): 2 \le x \le 5, 2 \le y \le 5\}, D = \{(x,y): (1 \le x \le 3, 1 \le y \le 3) \lor (2 \le x \le 4, 2 \le y \le 4)\}
$
The reported results are based on a test set of $100$ patterns per class, which are selected via the same sampling schema used to generate the training set. 
We used a Gaussian kernel with fixed $\sigma$ set to $0.4$ and averaged the accuracy results over $10$ different runs performed on different instances of the training and test sets.

\begin{figure}[t]
\centering
\vspace{-1cm}
\includegraphics[width=0.7\textwidth]{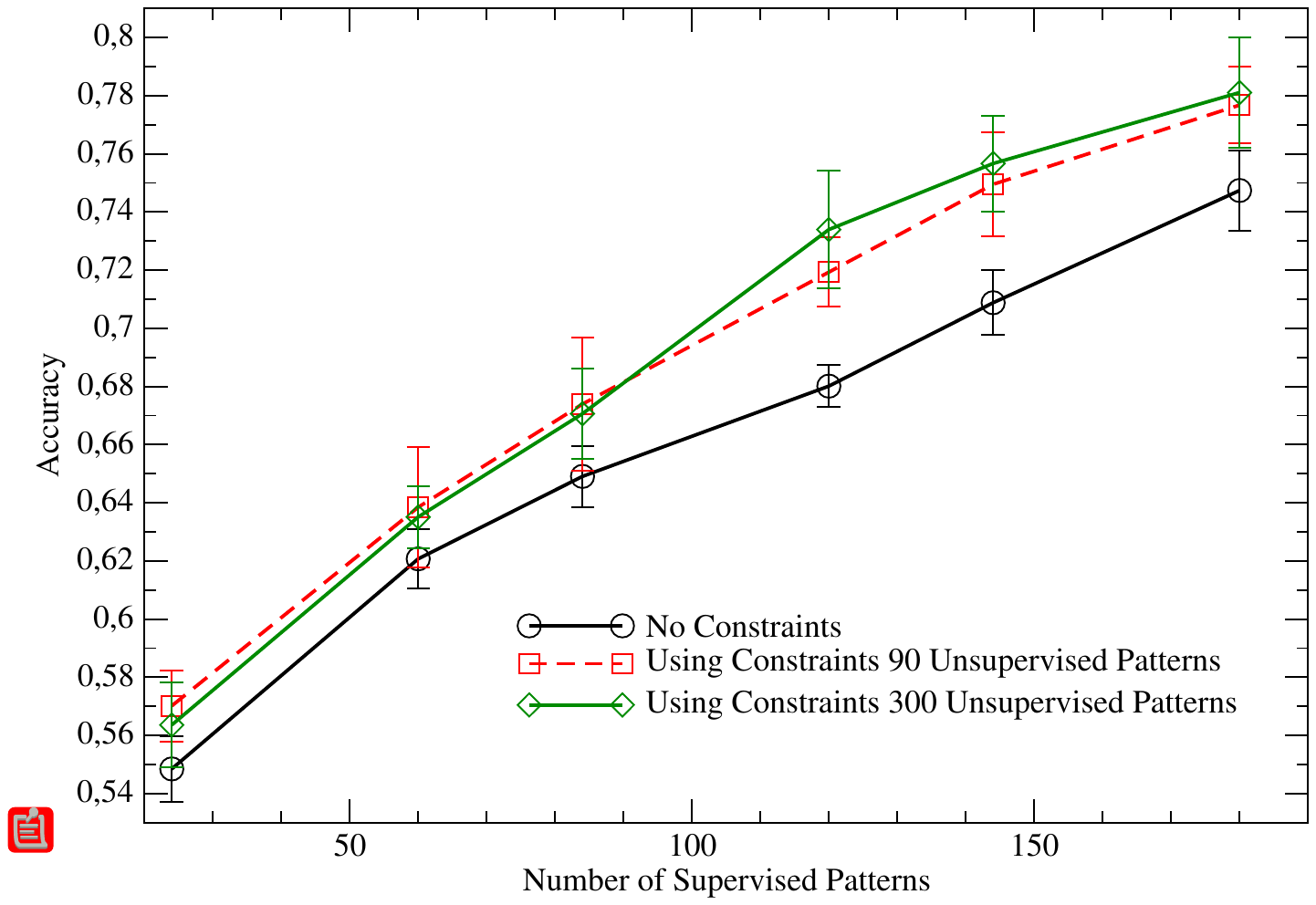}
\vspace{-0.6cm}
\caption{\scriptsize{Benchmark 1: classification accuracy for different labeled and unlabeled datasets when using or not using the constraints in training.}}
\label{fig:leoexample}
\vspace{-0.4cm}
\end{figure}
Let $a({\bf x}),b({\bf x}),c({\bf x}),d({\bf x})$ be a set of predicates representing whether a pattern ${\bf x}$ belongs to the classes $A,B,C,D$, respectively.
The following two FOL clauses are supposed to be known a-priori about the geometry of the classification task:
$\forall {\bf x} ~ \left( a({\bf x}) \land b({\bf x}) \right) \lor \left( b({\bf x}) \land c({\bf x}) \right) \Rightarrow d$ and $\forall {\bf x} ~ a({\bf x}) \lor b({\bf x}) \lor c({\bf x}) \lor d({\bf x})$.
Figure \ref{fig:leoexample} reports the classification accuracy obtained when using the constraints and the unsupervised data versus when no constraints are employed in learning the classification task. The classifier trained using the constraints outperforms the one learned without using the constraints by a statistically significant margin, which ranges between $2$\% and $5$\% depending on the training configuration.

In a second benchmark, we assume to have patterns laying in a $\RealSet^2$ feature space and belonging to two classes $A,B$, according to the well-known two moon-like shaped distributions.
We assume to be assigned an a-priori similarity relation $r({\bf x}, {\bf y})$ between a set of pairs of patterns $({\bf x}, {\bf y})$.
The semantic meaning of the relation $r$ can differ in different applications. For example, it could be used to represent the hyperlink connections between documents in Web retrieval tasks, or the co-citations among authors, etc.
In this experiment, we assume that $r$ models the geometric closeness of the patterns in the feature space. This assumption is very general and can be applied in any application where the input patterns lay in a metric space.
In particular, the following FOL clause is used to express the knowledge that the input patterns featuring a similarity relation should yield the same predicate output:
\begin{equation}
\forall {\bf x} \forall {\bf y} ~ r({\bf x}, {\bf y}) \Rightarrow (f({\bf x}) \land f({\bf y})) \lor (\lnot f({\bf x}) \land \lnot f({\bf y}))\ .
\label{eq:manifold_regularization}
\end{equation}
This is a reformulation of the well known assumption made in manifold regularization~\cite{belkin2006manifold} in a continuous logic setting.
This assumption expresses the fact that the input patterns are distributed along a manifold, over which the functions to be learned should be smooth, e.g. connected inputs on the manifold should tend to correspond to similar function outputs.
The FOL clause in equation~\ref{eq:manifold_regularization} can be rewritten as:
$
\forall {\bf x} \forall {\bf y} ~ \lnot (r({\bf x}, {\bf y}) \land \lnot (f({\bf x}) \land f({\bf y})) \land \lnot (\lnot f({\bf x}) \land \lnot f({\bf y})))\ .
$
\begin{figure}[t]
\centering
\vspace{-0.5cm}
\includegraphics[width=0.95\textwidth]{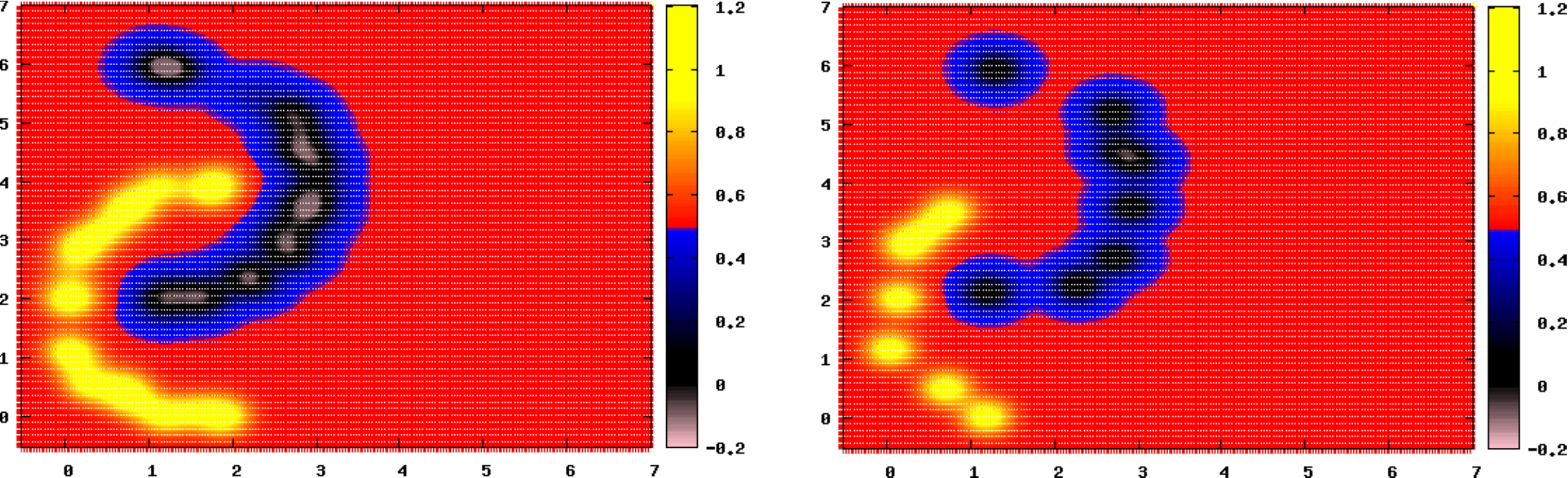}
\caption{\scriptsize{Predicate output when using $16$ labeled examples and using or not using the FOL clause on the left and right sides, respectively.}}
\label{fig:manifold_regularization}
\vspace{-0.4cm}
\end{figure}
Using the product t-norm and the mapping to a continuous cost function as explained in section~\ref{sec:learning_with_constraints}, we obtain the following constraint term for the cost function:
\begin{eqnarray*}
V(f) & = & \sum_{\bf x \in \mathcal{S}} \sum_{\bf y \in \mathcal{S}} ~ r({\bf x}, {\bf y}) ( 1 - f({\bf x}) f({\bf y})) (1 - (1 - f({\bf x})) (1 -f({\bf y}))) = \\
 & = &
\sum_{({\bf x}, {\bf y}): {\bf x},{\bf y}\in \mathcal{S}, r({\bf x}, {\bf y}) \ne 0} ~ r({\bf x}, {\bf y}) (1 - f({\bf x}) f({\bf y})) (1 - (1 - f({\bf x})) (1 -f({\bf y})))\ .
\end{eqnarray*}
In our experimental setting, the strength of the relation is computed as $r({\bf x}, {\bf y}) = e^{-||{\bf x} - {\bf y}||/\sigma_d}$, where $\sigma_d=\frac{2}{3}$.
The constraint part is then plugged into equation~\ref{eq:objective} and optimized by gradient descent.
Figure~\ref{fig:manifold_regularization} plots the output map of the learned predicate $f$. The effect of the knowledge expressed by the FOL clause over the unsupervised data smoothes the predicate output value over the regions where scarce labeled data is available. The activation map perfectly reconstructs the boundaries of the regions where the input patterns are distributed for the two classes.
\begin{table}[t]
\centering
\begin{tabular}{@{}lcccc@{}}
  & & \multicolumn{3}{c}{num labeled patterns} \\
  & & 4 & 8 & 12 \\
\hline
with FOL knowledge    & & 59.6\% & 68.5\% & 72.3\% \\
without FOL knowledge & & 40.4\% & 53.5\% & 71.2\% \\
\hline
\end{tabular}
\vspace{0.1cm}
\caption{\scriptsize{Moon benchmark: classification accuracy on the test set obtained with and without using the manifold regularization expressed in FOL form.}}
\label{tab:manifold_regularization}
\vspace{-0.8cm}
\end{table}
Table~\ref{tab:manifold_regularization} reports the accuracy values for different numbers of the labeled and unlabeled patterns. When learning using the FOL prior knowledge, the training data was augmented with $100$ unlabeled patterns. The accuracy values have been obtained as an average over $10$ different random generations of the training and test data. The accuracy gain is very significant when little labeled data is available.

\section{Conclusions}
\label{sec:conclusions}
This paper presents a framework to learn a set of predicates, each implemented as a kernel machine, starting from a collection of supervised training data and prior knowledge in form of FOL clauses.
The FOL clauses can be compiled into a set of real-valued function constraints that are subsequently converted to a penalty function added to the classic loss and regularization term of kernel machines. 
The experimental results show how it is possible to impose smoothness over a manifold of points using the FOL formalism. The regularized fitting of the supervised data along with the penalty term that expresses the logic clauses is optimized by a two-stage process, inspired to development psychology, that  is shown to be very effective especially when small sets of supervised data are available.
The proposed framework opens the doors to a new class of {\em semantic-based regularization machines}, in which it is possible to integrate prior knowledge using high level abstract representations.

\bibliography{regularization}
\bibliographystyle{splncs03}

\end{document}